\documentclass[10pt,twocolumn,letterpaper]{article}

\usepackage{iccv}
\usepackage{times}
\usepackage{epsfig}
\usepackage{graphicx}
\usepackage{amsmath}
\usepackage{amssymb}
\usepackage{graphicx}
\usepackage{amsmath}
\usepackage{amssymb}
\usepackage{booktabs}
\usepackage{times}
\usepackage{epsfig}
\usepackage{comment}
\usepackage{amsmath}
\usepackage{amssymb}
\usepackage{multirow}
\usepackage{booktabs}
\usepackage{graphicx}
\usepackage{pifont}
\usepackage{colortbl}
\usepackage[table]{xcolor}
\usepackage[ruled]{algorithm2e}
\definecolor{grey}{rgb}{0.89, 0.89, 0.89}

\usepackage[pagebackref=true,breaklinks=true,letterpaper=true,colorlinks,bookmarks=false]{hyperref}

\iccvfinalcopy 

\def\iccvPaperID{6347} 

\ificcvfinal\fi

\begin{document}

\title{Towards Open-Vocabulary Video Instance Segmentation}
\vspace{-4mm}
\author{
\hspace{3mm}Haochen Wang\footnotemark[1]\\
\hspace{3mm}University of Amsterdam\\
\hspace{3mm}{\tt\small h.wang3@uva.nl}
\and 
\hspace{20mm} Cilin Yan\footnotemark[1]\\
\hspace{20mm}Beihang University\\
{\hspace{20mm} \tt\small clyan@buaa.edu.cn}
\and
\hspace{8mm}Shuai Wang\\
\hspace{8mm}University of Amsterdam\\
\hspace{8mm}{\tt\small s.wang3@uva.nl}
\and
\hspace{-5mm}Xiaolong Jiang, Xu Tang, Yao Hu\\
\hspace{-5mm}Xiaohongshu Inc.\\
{\hspace{-5mm}\tt\small laige@xiaohongshu.com}
\and 
Weidi Xie\footnotemark[2]\\
Shanghai Jiao Tong University\\
{\tt\small weidi@robots.ox.ac.uk}
\and
Efstratios Gavves\\
University of Amsterdam\\
{\tt\small egavves@uva.nl}
}
\maketitle

\renewcommand{\thefootnote}{\fnsymbol{footnote}} 
\footnotetext[1]{Equal contribution.}  
\footnotetext[2]{Corresponding author.} 

\ificcvfinal\thispagestyle{empty}\fi

\begin{abstract}
Video Instance Segmentation~(VIS) aims at segmenting and categorizing objects in videos from a closed set of training categories, lacking the generalization ability to handle novel categories in real-world videos.
To address this limitation, we make the following three contributions.
First, we introduce the novel task of Open-Vocabulary Video Instance Segmentation, which aims to simultaneously segment, track, and classify objects in videos from open-set categories, including novel categories unseen during training.
Second, to benchmark Open-Vocabulary VIS, we collect a Large-Vocabulary Video Instance Segmentation dataset~(LV-VIS), that contains well-annotated objects from 1,196 diverse categories, significantly surpassing the category size of existing datasets by more than one order of magnitude.
Third, we propose an efficient Memory-Induced Transformer architecture, OV2Seg, to first achieve Open-Vocabulary VIS in an end-to-end manner with near real-time inference speed.
Extensive experiments on LV-VIS and four existing VIS datasets demonstrate the strong zero-shot generalization ability of OV2Seg on novel categories.
The dataset and code are released here \textcolor[rgb]{0,0,1}{https://github.com/haochenheheda/LVVIS}. 
\vspace{-4mm}
\end{abstract}

\begin{figure}[t]
\begin{center}
\vspace{-0mm}
\includegraphics[width=1\linewidth]{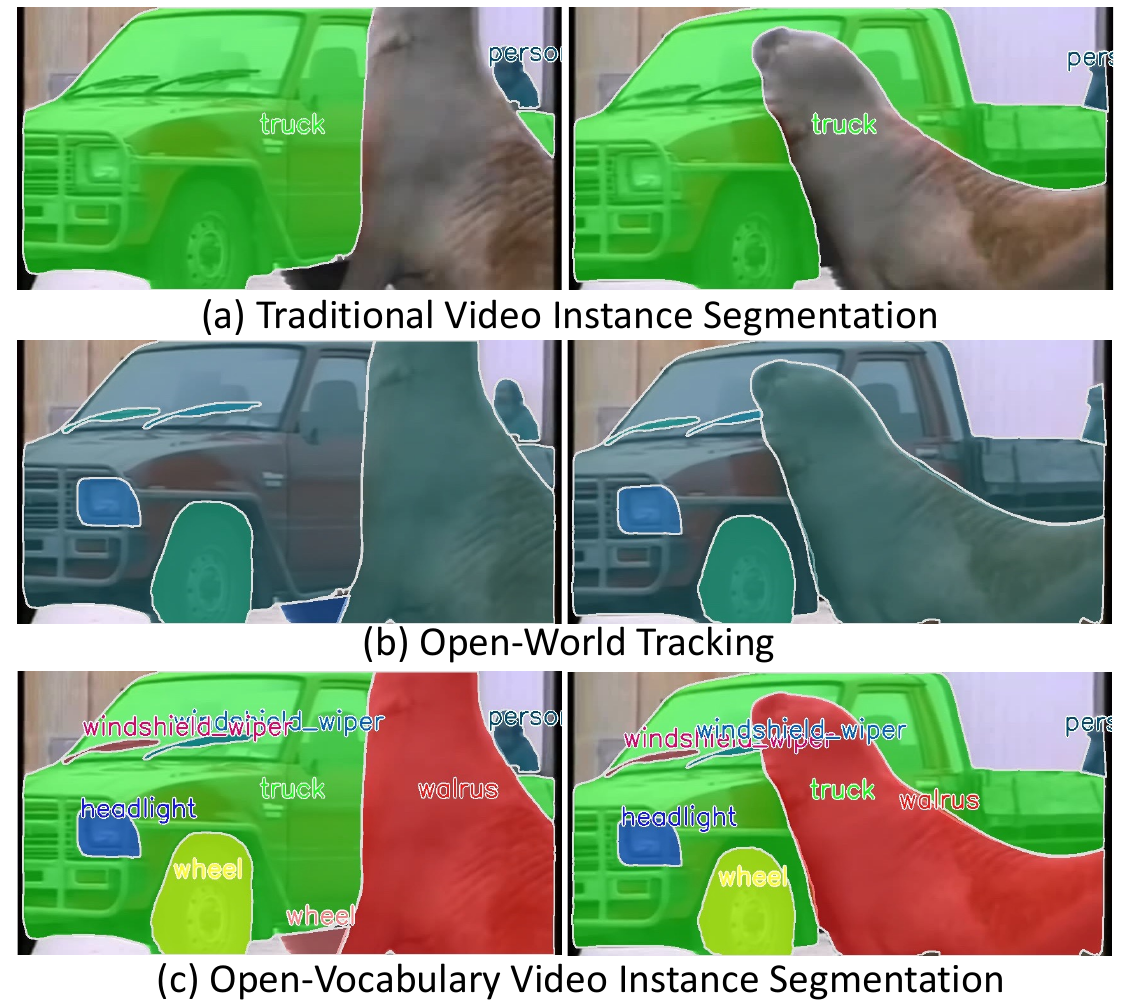}
\end{center}
\vspace{-2mm}
\caption{
(a) Traditional VIS fails to tackle objects from novel categories (unseen during training, e.g. walrus in the figure); (b) Open-World Tracking aims to segment and track all objects in a class-agnostic manner, while lacking the ability for object categorization in videos; (c) Open-Vocabulary VIS aims to simultaneously segment, track, and classify objects for both training categories, e.g., person, and novel categories, e.g., walrus in the figure. Different colors in the figures indicate different object instances.\vspace{-5mm}}
\label{fig1}
\end{figure}

\section{Introduction}
Despite impressive efforts, Video Instance Segmentation~(VIS)~\cite{yang2019video,cheng2021mask2former, wang2021end} is fundamentally constrained to segment and classify objects from a closed set of training categories, thus limiting the capacity to generalize to novel concepts, see Fig.~\ref{fig1}~(a).
In real-world scenarios, this closed-vocabulary paradigm lacks practical value, as the model usually encounters objects from novel categories unseen in the training stage.
In contrast, recent works Open-World Tracking~(OWTB)\cite{liu2022opening}, UVO~\cite{wang2021unidentified}, 
and BURST~\cite{athar2022burst} aim to segment and track all visible objects, see Fig.~\ref{fig1} (b).
However, those works still have no capacity to classify the objects from open-set categories, which is significant for video-level tasks, such as video captioning and action recognition.
To enhance the practicality of VIS, and make VIS more suited to downstream video-level tasks, \emph{we propose Open-Vocabulary Video Instance Segmentation} to simultaneously classify, track, and segment arbitrary objects from an open set of categories, as illustrated in Fig.~\ref{fig1} (c).

\begin{figure}[t]
\begin{center}
\vspace{-2mm}
\includegraphics[width=1\linewidth]{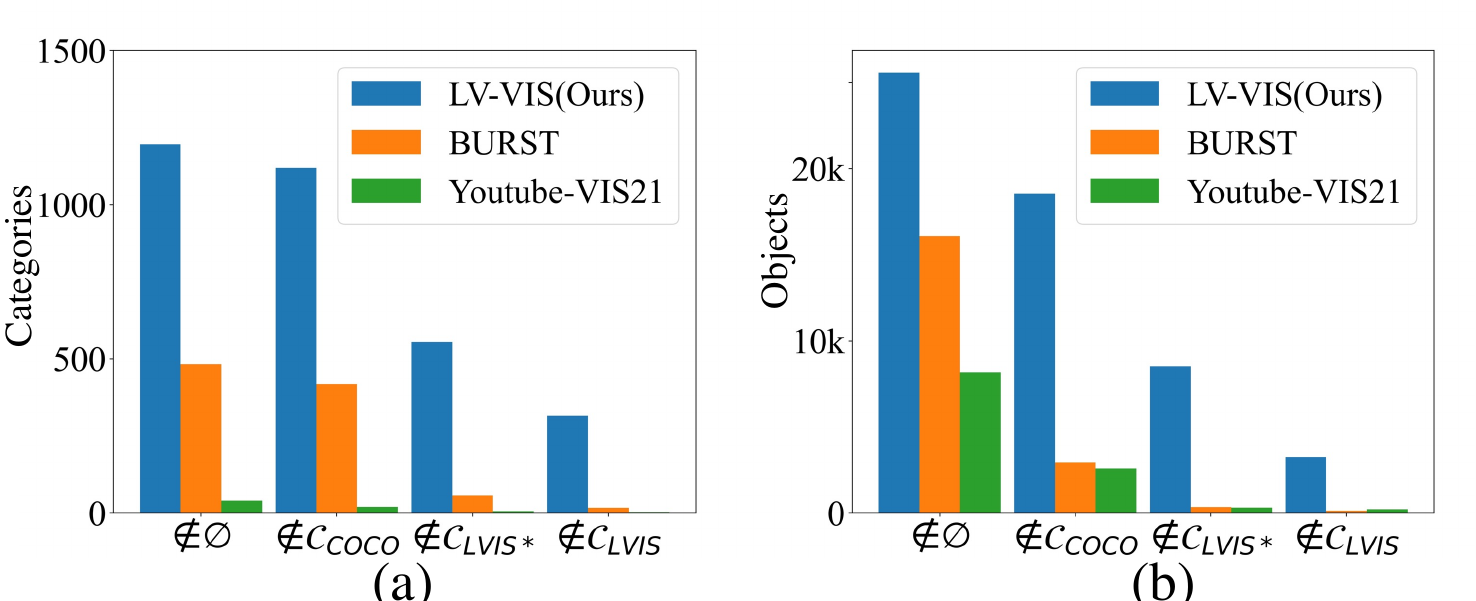}
\end{center}
\vspace{-4mm}
\caption{Figure (a) and (b) show the number of categories and object instances disjointed with a certain category set $\mathcal{C}$, respectively.
$\mathcal{C}_{\text{COCO}}$, $\mathcal{C}_{\text{LVIS}*}$, and $\mathcal{C}_{\text{LVIS}}$ represent all categories in MS-COCO, frequent/common categories in LVIS, and all categories in LVIS.
A larger percentage of categories and objects in LV-VIS are disjointed with the categories in commonly used datasets.
\vspace{-5mm}}
\label{fig2}
\end{figure}

To properly benchmark the Open-Vocabulary VIS, 
a video dataset with large and diverse object categories is necessary.
However, existing datasets Youtube-VIS~\cite{yang2019video}, OVIS~\cite{qi2021occluded}, and UVO~\cite{wang2021unidentified} are not sufficient as they contain only a few dozen of categories, see Tab.\ref{tab:statistics}.
BURST~\cite{athar2022burst} contains a relevant larger category set. 
Still,  81.7\% objects in BURST are from the common categories in MS-COCO, shown in Fig.~\ref{fig2} (b), and thus not diverse enough for the open-vocabulary evaluation of novel categories.

In this work, 
\emph{we collect a Large-Vocabulary Video Instance Segmentation dataset~(LV-VIS)}.
LV-VIS contains 4,828 videos and over 544k instance masks from 1,196 categories,
which is significantly larger than all existing video segmentation datasets.
Notably, as illustrated in Fig.~\ref{fig2} (b), a large percentage of annotated objects in LV-VIS are distinct from the categories in the commonly used datasets such as MS-COCO~\cite{lin2014microsoft} and LVIS~\cite{gupta2019lvis}, making it well suited for evaluating the generalization ability on novel categories of Open-Vocabulary VIS methods, and also a valuable supplement to the existing datasets.
\begin{table}[t]
\scalebox{0.77}{
\setlength\tabcolsep{0.1pt}
\centering
\begin{tabular}{c c c c c c c}
\toprule
Dataset&UVO~\cite{wang2021unidentified}&YT19~\cite{yang2019video}&YT21~\cite{yang2019video}& OVIS~\cite{qi2021occluded} & BURST~\cite{athar2022burst}&LV-VIS\\
\midrule
Videos&11228&2883 & 3859 & 901 & 2914 & 4828 \\
Instances&104898&4883 & 8171 & 5223 & 16089 & 25588 \\
Masks &593k& 131k& 232k & 296k & 600k & 544k \\
Mask/Frame&12.3&1.7&2.0&4.7&3.1&4.9\\
Object/Video&9.3&1.6&2.1&5.8&5.5&5.3\\
\rowcolor{grey} Categories&80$^*$&40 & 40 & 25 & 482 & 1196 \\
\bottomrule
\end{tabular}}
\vspace{3pt}
\caption{Key statistics comparison between our LV-VIS dataset and published video segmentation datasets. $*$ indicates the UVO only provides category labels for objects from 80 common categories defined in MS-COCO. Our LV-VIS dataset contains a significantly larger category set than the existing datasets.\vspace{-5mm}}
\label{tab:statistics}
\end{table}

Architecturally, a straightforward approach for Open-Vocabulary VIS is to associate per-frame results of open-vocabulary detectors~\cite{zhou2022detic, zang2022opendetr} with open-world trackers~\cite{liu2022opening}. 
However, this propose-reduce-association approach desires intricate hand-crafted modules such as non-maximum suppression, and neglects video-level features for stable tracking and open-vocabulary classification in videos, leading to sub-optimal performance and inference speed. 

In this work, we propose the \emph{first end-to-end Open-Vocabulary Video Instance Segmentation model}, OV2Seg, which simplifies the intricate propose-reduce-association paradigm and attains long-term awareness with a Memory-Induced Transformer.
Specifically, it starts by proposing and segmenting all objects with a Universal Object Proposal module, then a set of Memory Queries are introduced to incrementally encode object features through time, enabling long-term awareness for efficiently tracking all objects through time.
Lastly, given arbitrary category names as input, a language transformer encoder is adopted to classify the tracked objects based on the Memory Queries.
The Memory Queries incrementally aggregate the object features from different frames, thus leading to robust video object classification.
To our knowledge, OV2Seg is the first end-to-end model with the capability of segmenting, tracking, and classifying objects in videos from arbitrary open-set categories with near real-time inference speed.

We evaluate OV2Seg on LV-VIS 
as well as four existing video segmentation datasets: Youtube-VIS2021, Youtube-VIS2019, OVIS, and BURST.
\emph{Without finetuning on downstream video datasets}, OV2Seg matches the performance of several fully-supervised competitors~\cite{yang2019video, cao2020sipmask} that have been trained on target video datasets, while exhibiting strong zero-shot generalization ability for novel categories unseen during the training phase.

To sum up, our main contributions are:

(i) we introduce Open-Vocabulary VIS, which simultaneously segments, tracks, and classifies objects of arbitrary open-set categories in videos, generalizing the traditional VIS for closed-set training categories;

(ii) we collect a large-scale, pixel-level annotated video dataset LV-VIS, with a significantly larger vocabulary set than the existing video datasets, thus being a suitable testbed for Open-Vocabulary VIS;

(iii) we propose the first end-to-end Open-Vocabulary VIS model, OV2Seg, that can segment, track, and classify objects from novel categories with a Memory-Induced Transformer architecture.

\section{Related Work}

\noindent \textbf{Video Instance Segmentation} aims to segment, track and classify object instances from pre-defined training categories, such as people, cars, and dogs. 
Existing methods could be divided into frame-based methods~\cite{bertasius2020classifying,liu2021sg,lin2021video,yang2021crossover,he2017mask,yang2018efficient,han2022visolo} and clip-based methods~\cite{wang2021end,athar2020stem,hwang2021video,wu2021seqformer,cheng2021mask2former,yang2022temporally,heo2022vita}.
Frame-based methods, such as MaskTrack R-CNN~\cite{yang2019video}, propose to include an additional tracking head to the image instance segmentation model Mask R-CNN~\cite{he2017mask} for video instance segmentation.
The clip-based methods, such as IFS~\cite{hwang2021video}, SeqFormer~\cite{wu2021seqformer} and Mask2Former~\cite{cheng2021mask2former}, take the whole video or a clip of video as input and simultaneously output the masks and classification scores for all the input frames.
All the aforementioned models necessitate training on a substantial dataset of videos that include densely annotated objects from the training categories.
In contrast, our proposed OV2Seg is trained on image datasets, yet it can be directly applied to videos during testing and has the capability to tackle objects from novel categories.

\vspace{3pt}
\noindent \textbf{Open-Vocabulary Object Detection} aims to detect objects in images beyond a close vocabulary set, which is proposed in ZSD~\cite{bansal2018zero}.
Inspired by CLIP~\cite{radford2021learning}, ViLD~\cite{gu2021open} and ZSD-YOLO~\cite{xie2021zsd} propose to explicitly distill knowledge from the pre-trained CLIP visual embedding into the backbone of Faster RCNN.
Detic~\cite{zhou2022detic} and PromptDet~\cite{feng2022promptdet} strive to expand the vocabulary of the detector by self-training on additional images.
Those methods inherit the standard architecture of a two-stage detector, which follows a complex propose-and-reduce pipeline.
OV-DETR~\cite{zang2022opendetr} made the first attempt to model an end-to-end open vocabulary detector with an architecture like DETR~\cite{carion2020end}, while still limited in the inference speed caused by class-dependent object queries.

\vspace{3pt}
\noindent \textbf{Open World Tracking.}
Multiple Object Tracking~\cite{zhang2008global,tang2017multiple,braso2020learning} aims to track multiple objects through time.
OWTB~\cite{liu2022opening} proposed Open World Tracking to segment and track all the objects in videos.
OWTB achieves state-of-the-art performance on several multi-object tracking datasets. 
Recent work UVO~\cite{wang2021unidentified} focuses on class-agnostic object segmentation and tracking in videos, neglecting the object classification performance during evaluation.
Methods adopted in BURST~\cite{athar2022burst}, such as STCN~\cite{cheng2021rethinking} and MaskTrack~\cite{yang2019video}, have no ability to classify objects from novel categories.
Therefore BURST mainly follows the evaluation protocols of class-agnostic Multiple Object Tracking, 
while only measuring category-wise evaluation for the 80 training categories in MS-COCO.
On the contrary, we introduce Open-Vocabulary VIS to entail open-vocabulary object categorization along with segmentation and tracking in videos. 
We establish several baseline models for Open-Vocabulary VIS by integrating open-world tracking methods~\cite{liu2022opening} with open-vocabulary detectors~\cite{zhou2022detic, zang2022opendetr} to experimentally demonstrate the effectiveness of our proposed OV2Seg.

\begin{figure*}[!t]
\begin{center}
\includegraphics[width=0.98\linewidth]{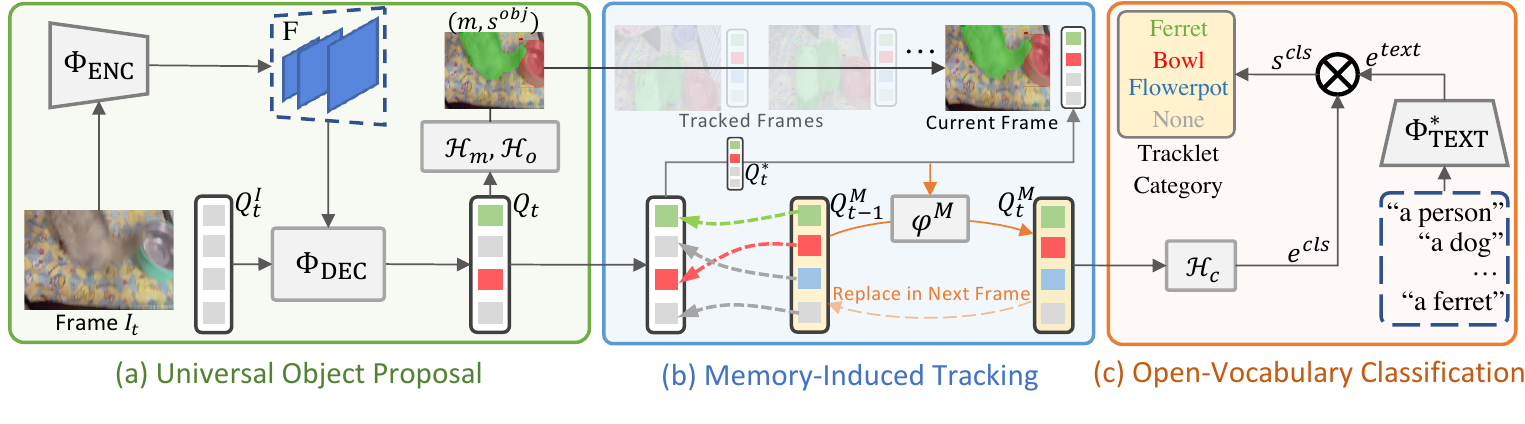}
\end{center}
   \vspace{-7mm}
\caption{Overview of OV2Seg. \textcolor[rgb]{0.35546875, 0.5546875, 0.1953125}{{(a) Universal Object Proposal}}: Input with frame $I_t$, a transformer encoder $\Phi_{\textsc{enc}}$ is adopted to extract multi-scale features $\mathcal{F}$. 
$N$ class-independent queries $Q^I_t \in \mathbb{R}^{N \times d}$ are fed to a transformer decoder $\Phi_{\textsc{dec}}$ to generate object-centric queries $Q_t$.
Then $Q_t$ is utilized to produce the segmentation mask $m \in \mathbb{R}^{N \times H \times W}$ and object score $s^{\text{obj}} \in \mathbb{R}^{N \times 1}$ with a mask head $\mathcal{H}_m$ and an object score head $\mathcal{H}_o$, respectively. 
\textcolor[rgb]{0.3203125, 0.484375, 0.80859375}{{(b) Memory-Induced Tracking}}: A set of Memory Queries $Q^M_{t-1} \in \mathbb{R}^{N \times d}$ is proposed to associate $Q_t$ with Hungarian Algorithm. $Q^M_{t-1}$ is updated by a function $\varphi^M$ to dynamically aggregate the associated object-centric queries $Q_t^*$ through time and obtain $Q^M_{t}$ for the tracking of the next frame.
\textcolor[rgb]{0.796875, 0.3984375, 0.0}{{(c) Open-Vocabulary Object Classification}}: A class head $\mathcal{H}_c$ is applied on $Q_t^M$ to generate the class embedding $\mathbf{e}^{\text{cls}} \in \mathbb{R}^{N \times d}$.
$\mathbf{e}^{\text{cls}}$ is then dot-producted with the text embedding $\mathbf{e}^{\text{text}}$ generated by a text encoder $\Phi_{\textsc{text}}\in \mathbb{R}^{|\mathcal{C}| \times d}$ to obtain the classification score $s^{cls}\in \mathbb{R}^{N \times |\mathcal{C}|}$ for a open set of categories $\mathcal{C}$.
\vspace{-4mm}}
\label{fig2:architecture}
\end{figure*}

\section{Setting of Open-Vcabulary VIS}
\label{sec:problem_scenario}
Given a training dataset $\mathcal{D}_{train}$ containing pixel-level annotations for a set of training categories $\mathcal{C}_{\text{train}}$, traditional VIS aims to train a model $f_{\theta}(\cdot)$ which could be tested on a dataset $\mathcal{D}_{\text{test}} = \{V_i\}_{i=1}^N$,
where $V_i \in \mathbb{R}^{T_i \times H_i \times W_i \times 3}$ refers to a video clip with shape of $(H_i, W_i)$ and $T_i$ frames.
$f_{\theta}(\cdot)$ is supposed to predict a sequence of segmentation mask $\{m_t\}_{t=1}^{T_i} \in \mathbb{R}^{T_i \times H_i \times W_i}$ and a category label $c \in \mathcal{C}_{train}$ for each object in videos from the training categories, while the objects from novel categories $\mathcal{C}_{novel}$ are ignored.

In contrast, Open-Vocabulary VIS aims to train a model $f_{\theta}(\cdot)$ on $\mathcal{D}_{train}$, and then test on $\mathcal{D}_{test}$ for both training categories $\mathcal{C}_{train}$ and novel categories $\mathcal{C}_{novel}$. 
Specifically, given a test video clip $V_i \in \mathbb{R}^{T_i \times H_i \times W_i \times 3}$ during inference, the trained model $f_{\theta}(\cdot)$ is supposed to predict the segmentation mask sequence $\{m_t\}_{t=1}^{T_i} \in \mathbb{R}^{T_i \times H_i \times W_i}$ and the category label $c \in (\mathcal{C}_{train} \cup \mathcal{C}_{novel})$ for each object $p$ in $V_i$:\vspace{-2mm}
\begin{align}
    \{\{m_1, m_2, ..., m_{T_i}\}, c\}_p^{P} = f_{\theta}(V_i),
\end{align}
where $P$ is the total number of objects in video $V_i$, class label $c$ belongs to the union of training and novel categories, and $m_t \in \mathbb{R}^{H_i \times W_i}$ is the segmentation mask for each frame $t$.
In the experiments, we term the training categories as base categories, while the categories disjointed with base categories as novel categories.

\vspace{3pt}
\noindent \textbf{Evaluation Metrics.}
We adopt the mean average precision (mAP) metric for both base categories and novel categories.
The mAP in videos is defined in Youtube-VIS~\cite{yang2019video}.
Specifically, we first compute the average precision~(AP) of each object category among all the test videos, defined as the area under the precision-recall curve plotting with the category confidence score.
The value of AP is measured under 10 intersection-over-union (IoU) thresholds from 50\% to 95\% at step 5\%, following the standard setting in MS-COCO~\cite{lin2014microsoft}.
The IoU in videos is defined in Youtube-VIS~\cite{yang2019video}, which computes the spatial-temporal consistency of predicted and ground truth segmentation masks.
Finally, the mean average precision~(mAP) is computed by averaging the AP over the base category set and novel category set, termed as mAP$_{b}$ and mAP$_{n}$.

\section{LV-VIS Dataset}
To properly evaluate the Open-Vocabulary VIS, we establish a new dataset named LV-VIS with a large vocabulary set. This section outlines the data collection and annotation procedure, 
and provides a comparison of key dataset statistics with existing video-level datasets.

\vspace{3pt}
\noindent \textbf{Data Collection and Manual Annotation.}
We start with collecting 7,569 video clips through the Internet by querying 1,621 well-defined category names.
Then we manually remove inappropriate video clips such as videos with sensitive personal information, and videos with too many objects to annotate.
Then we keep 4,828 videos in the LV-VIS.
We maintain the original resolution of the videos, varying from 480p to 1080p.
All the videos are pre-processed to a length between 2 and 10 seconds, and we extract 5 frames per second for all the processed videos.
We thoroughly annotate all clearly visible objects in the selected videos.
Each annotated object is assigned a category label, instance identity, and mask sequence in polygon format.
Then we remove the uncovered categories, merge categories with synonyms, and then retain 1,196 well-defined object categories.
The details are shown in Supplemental Materials Sec.~\textcolor[rgb]{1,0,0}{A}.
We provide the visualization of annotated video clips in Supplemental Materials Sec.~\textcolor[rgb]{1,0,0}{B}.

\vspace{3pt}
\noindent \textbf{Dataset Statistics.}
The LV-VIS dataset comprises 4,828 real-world videos with a total length of 6.2 hours and contains 544,451 pixel-level annotated segmentation masks from 1,196 categories. 
The categories in LV-VIS are split into 641 base categories~(seen during training) inherited from frequent and common categories in LVIS~\cite{gupta2019lvis}, and 555 novel categories disjoint with the base categories.
As shown in Tab. ~\ref{tab:statistics}, the vocabulary size of LV-VIS is significantly larger than
Youtube-VIS~(40 categories), OVIS~(25 categories), UVO~(80 categories), and BURST~(482 categories) respectively.
Moreover, 93.5\%/46.4\% of the categories in LV-VIS are disjoint from categories in MS-COCO and frequent/common categories in LVIS, 
whereas the corresponding values for BURST and Youtube-VIS are much lower at 86.7\%/11.8\% and 50\%/15\%, respectively.
In the meantime, 33.3\% objects in LV-VIS are from novel categories, while the BURST only contains 2.2\% objects from novel categories, see Fig.~\ref{fig2}.
That means LV-VIS contains a greater number of novel objects not shown in commonly used datasets such as MS-COCO and LVIS, making it a valuable supplement to the existing datasets and thus more suitable for the category-wise evaluation of novel categories.
Due to the space limitation, more extensive dataset statistics are shown in Supplementary Material Sec.~\textcolor[rgb]{1,0,0}{A}.

\vspace{3pt}
\noindent \textbf{Dataset Split.}
LV-VIS is split into a train set, a validation set, and a test set, which contains 3083, 837, and 908 videos respectively.
We will publicly release the videos and annotations of the train set and validation set,
and host an online evaluation server to benchmark related efforts with the challenging test set.

%


\section{Architecture of OV2Seg}
After constructing the dataset, we here present an end-to-end Open-Vocabulary VIS architecture, OV2Seg, as shown in Fig.~\ref{fig2:architecture}. 
Generally speaking, the proposed architecture consists of three modules, 
(i) a Universal Object Proposal module, proposes and segments objects with a set of class-independent object queries;
(ii) a Memory-Induced Tracking module, tracks the object proposals from consecutive frames with a set of Memory Queries, which incrementally aggregate object features through time to establish long-term awareness;
(iii) an Open-Vocabulary Classification module, classifies the tracked objects with the text embeddings computed from a pre-trained text encoder given arbitrary category names as input.
We detail the architecture in the following sections.

\subsection{Universal Object Proposal}
We feed each input frame $I_t$ into a transformer encoder~
$\Phi_{\textsc{enc}}$, consisting of multiple deformable attention layers~\cite{zhu2020deformable}, 
to get the multi-scale feature maps $\mathcal{F}$:
\begin{align}
    &\mathcal{F} = \{ f^1, \dots, f^l\}=  \Phi_{\textsc{enc}}(I_t)
\end{align}
where $f^l$ refers to the feature map at $l$-th level. 
To propose objects from all categories, one state-of-the-art image-level open-vocabulary detector OV-DETR~\cite{zang2022opendetr} uses class-dependent object queries conditioned on the given category names.
In this way, each category is processed by a set of separate object queries, leading to an infeasible inference speed on real-world videos.

In this work, we experimentally prove that a set of class-independent object queries could efficiently propose objects from all categories, avoiding separate processes for each category.
Therefore, we adopt $N$ class-independent learnable object queries $Q^I \in \mathbb{R}^{N \times d}$ to obtain the class-agnostic object proposals.
Specifically, the object queries $Q^I$ and the multi-scale feature maps $\mathcal{F}$ are fed into a transformer decoder  $\Phi_{\textsc{dec}}$, where the $Q^I$ iteratively attend to each layer of $\mathcal{F}$, generating object-centric queries $Q$:\vspace{-2mm}
\begin{align}
  Q& = \Phi_{\textsc{dec}}(\mathcal{F}, Q^I) \in \mathbb{R}^{N \times d}.
\end{align}

Then the output object-centric queries $Q$ from the transformer decoder are passed into a mask generation head $\mathcal{H}_{m}$ and an object score head $\mathcal{H}_{o}$:\vspace{-0.5mm}
\begin{align}
  m, s^{\text{obj}} & =  \mathcal{H}_m(Q) \circledast f^{-1}, \mathcal{H}_o(Q),
\end{align}
\vspace{-0.5mm}
where $f^{-1}$ is the last layer of $\mathcal{F}$.
$\mathcal{H}_{m}(\cdot)$ consists of 3 multilayer perceptron (MLP) layers and outputs the dynamic convolutional kernel for each object query.
The dynamic convolution kernels are then convolved with $f^{-1}$ to obtain the per-frame segmentation masks $m \in \mathbb{R}^{N \times H \times W}$.
$\mathcal{H}_{o}(\cdot)$ also consists of 3 MLP layers and outputs binary scores $s^{\text{obj}}\in \mathbb{R}^{N \times 1}$, 
indicating if a query represents an object.


\subsection{Memory-Induced Tracking}
The object-centric queries $Q$ in consecutive frames could be associated with the Hungarian algorithm to generate the object tracklets in videos, inspired by MinVIS~\cite{huang2022minvis}.
However, directly associating objects of consecutive frames neglects
 long-term dependencies in videos.
As such, this approach fails in complex cases such as objects disappearing and reappearing in the videos.

To this end, we propose to maintain a set of Memory Queries $Q^M \in \mathbb{R}^{N \times d}$ to obtain long-term dependency, see Fig.~\ref{fig2:architecture} (b). 
Specifically, we compute the inner-product similarity between Memory Queries $Q^M_{t-1}$ of the last frame and the object-centric queries $Q_t$.
Each object-centric query is associated with one of the Memory Queries by the Hungarian algorithm upon the similarity matrix.
The $Q^M_0$ is initialized by the $Q_0$ of the first frame.
Afterward, it is momentum updated through time with an update function $\varphi^{M}(\cdot, \cdot)$, thereby gradually encoding the video-level object features to tackle objects with disappear or serious occlusion.
Specifically, $\varphi^M(\cdot, \cdot)$ is defined as:\vspace{-0.5mm}
\begin{align}
Q^M_t &= \varphi^M( Q^M_{t-1}, Q_t^*)\\
   &= \alpha \cdot s^{\text{obj}} \cdot Q_t^* + (1 - \alpha \cdot s^{\text{obj}}) \cdot Q^M_{t-1},
\label{eqa:memory}
\end{align}
\vspace{-0.5mm}
where $\alpha$ is a factor to control the update ratio and $s^{\text{obj}}$ is the object score for each query.
The $Q_t^*$ is the associated object-centric queries after Hungarian algorithm.
If an object is seriously occluded or disappears during a video, the corresponding object score $s^{\text{obj}}$ tends to be low and therefore constrains the memory update process to keep the object memory features unchanged.
By maintaining a small set of Memory Queries, OV2Seg efficiently tracks all the segmented objects through time with long-term awareness, and enhances the following object classification with incremental video object features.

\subsection{Open-Vocabulary Classification}
We feed the vocabulary set into a frozen pretrained CLIP Text Encoder $\Phi_{\textsc{text}}$ to generate the text embedding as the classifiers.
For example, a `dog' classifier can be:
\begin{align}
    \mathbf{e}^{\text{text}} = \Phi_{\textsc{text}}(\text{``this is a photo of [dog]''}),
\end{align}
in which the vocabularies could be replaced by arbitrary novel categories during inference.

The class embeddings $\mathbf{e}^{\text{cls}} \in \mathbb{R}^{N \times d}$ for the tracked objects are obtained by $\mathbf{e}^{\text{cls}} = \mathcal{H}_c(Q_t^M)$, where $\mathcal{H}_{c}(\cdot)$ indicates the class head consisting three MLP layers.
Then we can obtain the classification score $s^{\text{cls}}$ for each object tracklet by computing the cosine similarity between class embedding $\mathbf{e}^{\text{cls}} \in \mathbb{R}^{N \times d}$ and text embedding $\mathbf{e}^{\text{text}} \in \mathbb{R}^{\mathcal{|C|} \times d}$:
\begin{align}
s^{\text{cls}}_{i,j} = \sigma(\text{cos}( \mathbf{e}^{\text{cls}}_{i},\mathbf{e}^{\text{text}}_{j})/\epsilon),
\end{align}
where $i \in [1,N]$ and $j \in [1, |\mathcal{C}|]$ are indexes of object query and text embedding. $\sigma$ is the sigmoid function.
$\text{cos}(\cdot, \cdot)$ refers to the cosine similarity, and $\epsilon$ is the temperature hyper-parameter.
The memory queries $Q^M_t$ aggregate the object features from all frames weighted by the object score, the object features with low confidence are constrained thus leading to robust video object classification.

\subsection{Training and Loss}
As there is no learnable parameter in the Memory-Induced Tracking, we can exclude the tracking module and train OV2Seg on image-level datasets, such as LVIS.
This reduces the training resource requirements of large-scale video datasets and improves training efficiency.
We adopt a similar training loss in DETR~\cite{carion2020end} for set prediction:
\begin{equation}
    \begin{split}
    \mathcal{L}_{\text{match}}(\hat{y},y) = &
    \lambda_{\text{obj}}\mathcal{L}_{\text{obj}}(\hat{s}^{\text{obj}},s^{\text{obj}}) + 
    \lambda_{\text{cls}}\mathcal{L}_{\text{cls}}(\hat{s}^{\text{cls}},s^{\text{cls}}) \\ 
    & + \lambda_{\text{mask}}\mathcal{L}_{\text{mask}}(\hat{m},m).
    \end{split}
\end{equation}

We use the binary cross-entropy loss for classification loss $\mathcal{L}_{\text{cls}}$ and object score loss $\mathcal{L}_{\text{obj}}$.
The mask loss $\mathcal{L}_{\text{mask}}$ is the sum of dice loss~\cite{milletari2016v} and binary focal loss.
The ground truth $\hat{y}$ consists of an object label $\hat{s}^{\text{obj}}$, a class label $\hat{s}^{\text{cls}}$ and segmentation mask $\hat{m}$ for each object.
We assign the ground truth set $\hat{y}$ to the prediction set $y$ by minimizing the total loss function $\mathcal{L}_{\text{match}}$.

\section{Experiments}
We first introduce the datasets and evaluation metrics.
Then we describe the method implementations. 
Lastly, we present a thorough analysis of the experimental results.

\subsection{Datasets and Metrics}
\label{sec:dataset}
We train OV2Seg on the union of common and frequent categories in LVIS~\cite{gupta2019lvis}, and evaluate the performance on multiple video instance segmentation datasets, including our collected LV-VIS dataset, as well as Youtube-VIS2019~\cite{yang2019video}, Youtube-VIS2021~\cite{yang2019video}, OVIS~\cite{qi2021occluded}, and BURST~\cite{athar2022burst}. 
Notably, we do not finetune OV2Seg on any videos from the target datasets.

\noindent \textbf{LVIS} is a widely used image open-vocabulary detection dataset, which contains a large set of 1,203 categories. 
Following ViLD~\cite{gu2021open}, we take frequent and common categories as the base categories and all categories disjointed with base categories as novel categories.

\noindent \textbf{Youtube-VIS} is the most widely used video instance segmentation dataset, which contains 40 categories for both YVIS19 and YVIS21 versions.
We divide the categories in YVIS19 and YVIS21 into the base and novel categories following the partitions in LVIS.

\noindent \textbf{BURST} is a recently published video dataset extending TAO~\cite{dave2020tao}.
BURST contains 425 base categories and 57 novel categories following the partitions in LVIS.

\noindent \textbf{OVIS} has 25 object categories, with only one category not in LVIS base categories; therefore, we only report the performance of the overall categories for OVIS.

\noindent \textbf{Metrics.} We report the mean Average Precision mAP for overall categories, and further break it down to mAP$_b$ for base categories and mAP$_n$ for novel categories. 
For the evaluation of LVIS~\cite{gupta2019lvis}, we report the mAP for overall categories and mAP$_{n}$ for novel~(rare) categories.

\subsection{Implementation Details}
\label{sec:impl}
\noindent \textbf{Baseline Models}.
We build several baseline models by combining off-the-shelf two-stage open-vocabulary detection methods with the open-world tracking methods following the propose-reduce-association scheme.
Specifically, we generate 300 object proposals for each frame by the open-vocabulary detectors~\cite{zhou2022detic, du2022learning}.
Then we feed the proposals to SORT~\cite{bewley2016simple} and OWTB~\cite{liu2022opening} to achieve object tracking.
The SORT is a classic multiple object tracking method, and the OWTB is one of the state-of-the-art open-world tracking methods, that combines the IOU matrix and ROI feature similarity to estimate proposal similarity across frames.
For efficient evaluation, we remove the optical flow in OWTB.
In this way, we build four baseline models, 
namely, Detic-SORT, Detic-OWTB, Detpro-SORT, and Detpro-OWTB, to compare with our proposed OV2Seg.
We further design a two-stage method by propagating the segmentation masks of Detic with a state-of-the-art semi-supervised video object segmentation method XMem~\cite{cheng2022xmem}. 

\vspace{3pt}
\noindent \textbf{OV2Seg}. 
We experiment with both ResNet50~\cite{he2016deep} and SwinB~\cite{liu2021swin} as backbones.
We adopt 6 multi-scale deformable attention layers~\cite{zhu2020deformable} as the transformer encoder and 6 masked-attention layers~\cite{cheng2021mask2former} as the transformer decoder.
We use the pretrained text encoder extracted from the CLIP ViT-32 model to generate the open-vocabulary classifiers.
Parameters in the text encoder are frozen during the training.
The temperature hyper-parameter $\epsilon$ is set to 50.
We adopt the prompt learning in Detpro~\cite{du2022learning} to replace the hand-crafted prompt in the text encoder module.
The number of object queries is set to 100 for all experiments.
We set $\lambda_{\text{obj}}$=2, $\lambda_{\text{cls}}$=2, $\lambda_{\text{mask}}$=5 as the weight of each loss.
The memory update factor $\alpha$ is set to 0.7.


\vspace{3pt}
\noindent \textbf{Training Details}. 
We train OV2Seg on LVIS for 30 epochs with a batch size of 8.
We adopt AdamW~\cite{loshchilov2018fixing} optimizer.
The base learning rate is set to 1e-4 and then reduced by a factor of 10 at 0.9 and 0.95 fractions of the total number of training steps.
The learning rate of backbone parameters is factorized by 0.1.
We use the large-scale jittering ~\cite{du2021simple} with a random scale sampled from the range 0.1 to 2.0.
For efficient training, 
we only sample 50 categories to compute the binary cross-entropy loss in one iteration.
The training is conducted on 4 A100 GPUS for around 36 hours and all the inferences are conducted on a single A100 GPU. 

\subsection{Results}
\label{sec:results}

In this section, we first report the result comparison on the LV-VIS dataset between OV2Seg and a set of strong baseline models.
Then we conduct extensive ablation studies to illustrate the effectiveness of the main modules in OV2Seg.
Finally, we report the zero-shot generalization performance of OV2Seg on existing video instance segmentation datasets: Youtube-VIS2019, Youtube-VIS2021, BURST, and OVIS.

\subsubsection{Results on LV-VIS dataset}
We compare our OV2Seg with the propose-reduce-association baseline models on the LV-VIS dataset in Tab.~\ref{tab:OVVIS}.
OV2Seg with ResNet-50 backbone achieves 11.9 mAP$_n$ and 8.9 mAP$_n$ on LV-VIS validation set and test set, outperforming the DetPro-XMem by 6.5 and 5.3.
As shown in Tab.~\ref{tab:lvis}, OV2Seg matches the performance of Detpro~\cite{du2022learning} and Detic~\cite{zhou2022detic} on image datasets, while outperforming Detpro-OWTB and Detic-OWTB by a large margin on video dataset LV-VIS.
The comparison outlines the effectiveness of OV2Seg on video-level tasks.
This is mainly because the several-stage baseline models require complex hand-craft association rules and classify the objects by averaging the per-frame classification scores, ignoring the video-level correspondence.

In contrast, OV2Seg associates the object queries with the Memory Queries, which encodes the long-term object features for robust tracking and open-vocabulary classification in videos.
Notably,  OV2Seg outperforms Detic-XMem by a relative 120\% improvement on novel categories of the LV-VIS validation set.
This is significantly higher than the 28.4\% improvement on base categories.
As such, we conclude that OV2Seg has generalization abilities on novel categories in real-world videos.
As shown in Tab.~\ref{tab:OVVIS}, the OV2Seg with Resnet50 backbone achieves 20.1 frame-per-second on LV-VIS, outperforming the Detic-OWTB by 14.2.
We provide some visualizations in Fig.~\ref{fig:main_vis}.
Due to space limitations, more visualizations and failure cases are depicted in the Supplementary Material Sec. \textcolor[rgb]{1,0,0}{C}.
\begin{table}[t]
\scalebox{0.78}{
\setlength\tabcolsep{2pt}
\centering
\begin{tabular}{c c c c c c c c  c}
\toprule
\multirow{2}{*}{Method}&\multirow{2}{*}{Backbone}&\multicolumn{3}{c}{Val}&\multicolumn{3}{c}{Test}&\multirow{2}{*}{fps}\\
\cmidrule(lr){3-5}
\cmidrule(lr){6-8}
&&AP&AP$_{b}$&AP$_{n}$&AP&AP$_{b}$&AP$_{n}$&\\
\midrule
DetPro~\cite{du2022learning}-SORT~\cite{bewley2016simple}&R50&6.4&10.3&3.5&5.8&10.8&2.1&3.1\\
Detic~\cite{zhou2022detic}-SORT~\cite{bewley2016simple}&R50&6.5&10.7&3.4&5.7&10.6&2.1&6.0\\
DetPro~\cite{du2022learning}-OWTB~\cite{liu2022opening}&R50&7.9&12.9&4.2&7.0&12.6&2.9&3.1\\
Detic~\cite{zhou2022detic}-OWTB~\cite{liu2022opening}&R50&7.7&12.6&4.2&7.0&12.8&2.8&5.9\\
Detic~\cite{zhou2022detic}-XMem~\cite{cheng2022xmem}&R50&8.8&13.4&5.4&7.7&13.3&3.6&16.4\\
OV2Seg(Ours)&R50&\textbf{14.2}&\textbf{17.2}&\textbf{11.9}&\textbf{11.4}&\textbf{14.9}&\textbf{8.9}&\textbf{20.1}\\
\midrule
Detic~\cite{zhou2022detic}-SORT~\cite{bewley2016simple}&SwinB&12.8&21.1&6.6&9.4&15.8&4.7&6.7\\
Detic~\cite{zhou2022detic}-OWTB~\cite{liu2022opening}&SwinB&14.5&22.6&8.5&11.8&19.6&6.1&6.8\\
Detic~\cite{zhou2022detic}-XMem~\cite{cheng2022xmem}&SwinB&16.3&24.1&10.6&13.1&20.5&7.7&13.4\\
OV2Seg(Ours)&SwinB&\textbf{21.1}&\textbf{27.5}&\textbf{16.3}&\textbf{16.4}&\textbf{23.3}&\textbf{11.5}&\textbf{16.8}\\
\bottomrule
\end{tabular}
}
\vspace{-3pt}
\caption{The performance comparison on LV-VIS validation and test set. The AP, AP$_b$, and AP$_n$ mean the average precision of overall categories, base categories, and novel categories.\vspace{-3mm}}
\label{tab:OVVIS}
\end{table}

\begin{figure}[t]
\begin{center}
\includegraphics[width=1\linewidth]{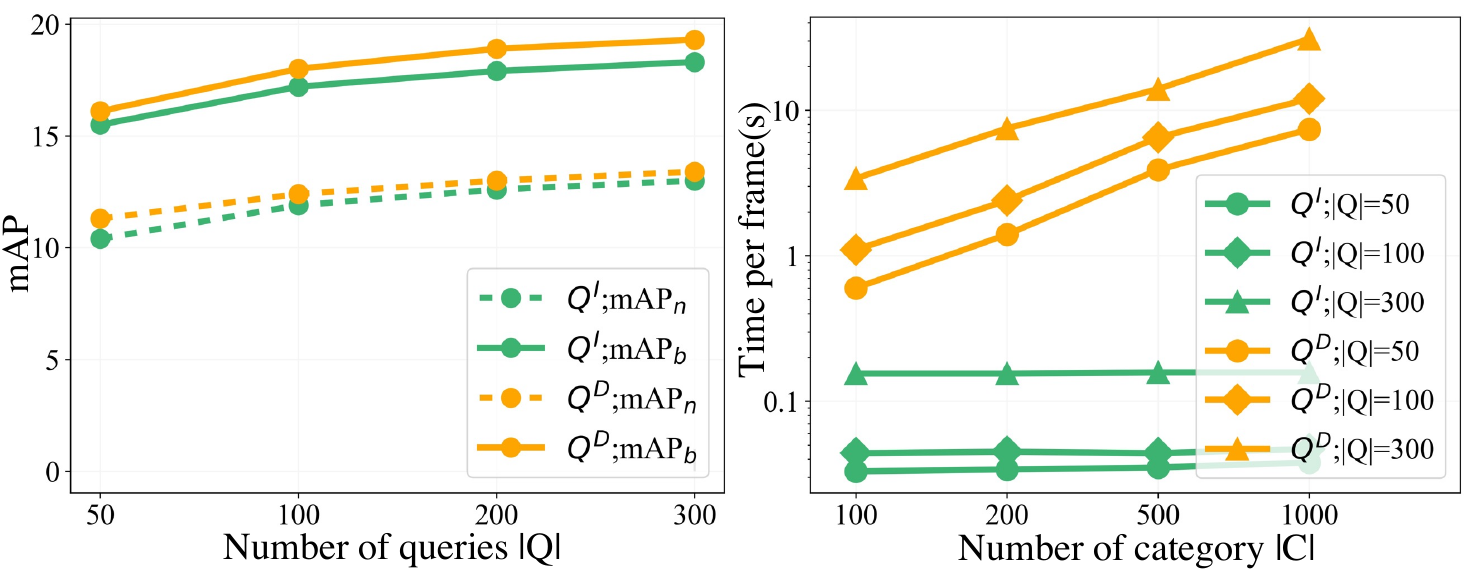}
\end{center}
   \vspace{-5mm}
\caption{mAP and inference time per frame~(s) comparison on LV-VIS validation dataset between class-dependent queries~($Q^D$) and class-independent queries~($Q^I$).}
\vspace{-4mm}
\label{fig:query_type}
\end{figure}

\subsubsection{Ablation Study}
\label{sec:ab}
We conduct extensive ablation studies on the LVIS, LV-VIS validation, and Youtube-VIS2019 datasets to thoroughly validate the effectiveness of the main modules in OV2Seg.
\vspace{3pt}

\noindent \textbf{Image-level Open-Vocabulary Detection.}
By removing the Memory-Induced Tracking module, OV2Seg can also conduct image-level Open-Vocabulary Detection.
We compare OV2Seg with several state-of-the-art open-vocabulary detection methods on LVIS and LV-VIS(O)~(image oracle evaluation) to demonstrate the effectiveness and efficiency of OV2Seg on image-level tasks.
As shown in Tab.~\ref{tab:lvis}, OV2Seg achieves 17.5 and 7.4 mAP$_n$ on LVIS and LV-VIS(O) respectively with Resnet50 backbone.
This matches existing two-stage state-of-the-art approaches Detic~\cite{zhou2022detic} and Detpro~\cite{du2022learning}, while allowing for faster inference speeds.
Specifically, Detic and Detpro require separate region proposal networks and box regression modules to generate the objects, as well as non-maximum suppression to reduce the redundant objects, resulting in a slow inference speed of 6.2 fps and 3.4 fps.
OV-DETR~\cite{zang2022opendetr}, the first transformer-based open-vocabulary detector, achieves 17.4 and 7.2 mAP$_n$ on LVIS and LV-VIS(O) with inference speed of 0.1 fps, limited by the class-dependent object queries.
In contrast, our proposed OV2Seg uses class-independent object queries, avoiding separate processes for each category.
As a result, OV2Seg achieves 21.6 fps, outperforming the fps of OV-DETR for more than two orders of magnitude.
More generally, we claim that high accuracy on image datasets is not the main goal of this work.
Instead, we prefer a simple and effective model for Open-Vocabulary VIS.
Therefore, we do not apply self-training and information distillation in our work, which are adopted in Detic and Detpro. 

\begin{table}[t]
\footnotesize
\setlength\tabcolsep{3.5pt}
\centering
\begin{tabular}{c c c c c c c c}
\toprule
\multirow{2}{*}{Method}&\multirow{2}{*}{Backbone}&\multirow{2}{*}{Stage}&\multicolumn{2}{c}{LVIS}& \multicolumn{2}{c}{LV-VIS(O)}& \multirow{2}{*}{fps}\\
\cmidrule(lr){4-5}
\cmidrule(lr){6-7}
&&&AP&AP$_n$& AP&AP$_n$&\\
\midrule
ViLD~\cite{gu2021open}&R50&2&25.5&16.6&-&-&-\\
DetPro$^*$~\cite{du2022learning}&R50&2&25.7&18.7&10.8&7.6&3.4\\
Detic~\cite{zhou2022detic}&R50&2&30.2&16.4&11.0&6.9&6.2\\
OV-DETR~\cite{zang2022opendetr}&R50&1&26.6&17.4&11.4&7.2&0.10\\
OV2Seg(Ours)&R50&1&25.4&17.5&11.0&7.4&21.6\\
\midrule
Detic~\cite{zhou2022detic}&SwinB&2&38.4&21.9&13.9&7.8&7.6\\
OV2Seg(Ours)&SwinB&1&31.4&21.8&14.2&8.3&18.2\\
\bottomrule
\end{tabular}
\vspace{3pt}
\caption{Image-level performance on LVIS dataset and LV-VIS validation dataset. The LV-VIS(O) means the oracle image evaluation, where we regard all the frames in LV-VIS as an image dataset and evaluate on the images separately. DetPro$^*$ is conducted by single-scale inference for fair comparison. \vspace{-2mm}}
\label{tab:lvis}
\end{table}

\begin{figure}[t]
\begin{center}
\vspace{-2mm}
\includegraphics[width=1\linewidth]{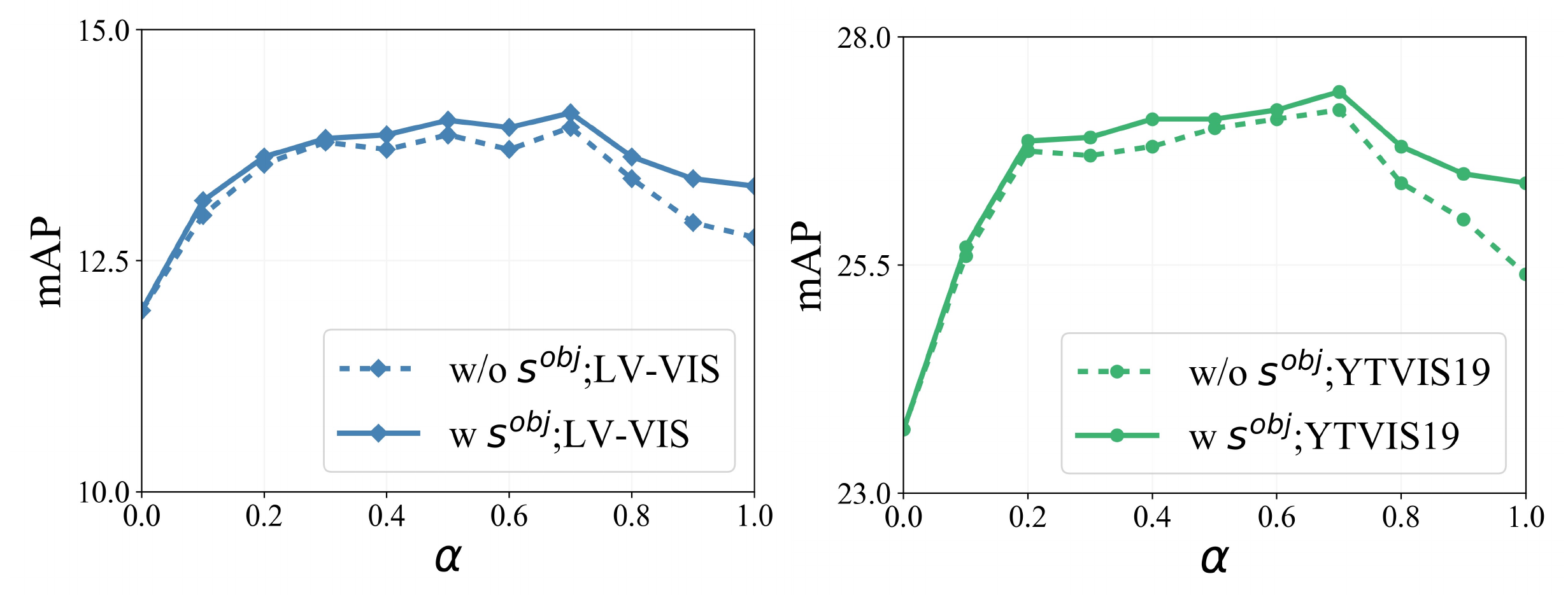}
\end{center}
   \vspace{-5mm}
\caption{Overall mean Average Precision~(mAP) under different memory update factor $\alpha$ on LV-VIS and Youtube-VIS2019 validation datasets. The w/o $s^{\text{obj}}$ means removing the object score $s^{\text{obj}}$ in the memory update module. }
\vspace{-5mm}
\label{fig:memory_update}
\end{figure}

\begin{table*}[t!]
\centering
\setlength\tabcolsep{7.4pt}
\scalebox{0.85}{
\begin{tabular}{c c c c c c c c c c c c c}
\toprule
\multirow{2}{*}{Method}&\multirow{2}{*}{Open}&\multirow{2}{*}{Backbone}&\multicolumn{3}{c}{YTVIS2019}&\multicolumn{3}{c}{YTVIS2021}&\multicolumn{3}{c}{BURST}&OVIS\\
\cmidrule(lr){4-6}
\cmidrule(lr){7-9}
\cmidrule(lr){10-12}
\cmidrule(lr){13-13}
&&&mAP&mAP$_b$&mAP$_n$&mAP&mAP$_b$&mAP$_n$&mAP&mAP$_b$&mAP$_n$&mAP\\
\midrule
FEELVOS~\cite{yang2019video} &\ding{55} & R50 & 26.9 &- &- &-&-&-&-&-&-&9.6 \\
MaskTrack~\cite{yang2019video} &\ding{55} & R50 & 30.3 &- &- &28.6&-&-&-&-&-&10.8\\
SipMask~\cite{cao2020sipmask} &\ding{55} &R50 & 33.7 & - & - & 31.7 & - & - & - & -&-  & 10.2 \\
Mask2Former~\cite{cheng2021mask2former} &\ding{55}& R50 & 46.4 & - & - & 40.6 &-&-&-&-&-&17.3\\
\midrule
\midrule
Detic~\cite{zhou2022detic}-SORT~\cite{bewley2016simple}&\ding{51}&R50&14.6&17.0&3.5&12.7&14.4&3.1&1.9&1.8&\textbf{2.5}&6.7\\
Detic~\cite{zhou2022detic}-OWTB~\cite{liu2022opening}&\ding{51}&R50&17.9&20.7&4.5&16.7&18.6&5.8&2.7&2.8&1.8&9.0\\
OV2Seg(Ours)&\ding{51}&R50&\textbf{27.2}&\textbf{30.1}&\textbf{11.1}&\textbf{23.6}&\textbf{26.5}&\textbf{7.3}&\textbf{3.7}&\textbf{3.9}&2.4&\textbf{11.2}\\
\midrule
Detic~\cite{zhou2022detic}-SORT~\cite{bewley2016simple}&\ding{51}&SwinB&23.8&27.2&7.9&21.6&23.7&9.8&2.5&2.7&1.0&11.7\\
Detic~\cite{zhou2022detic}-OWTB~\cite{liu2022opening}&\ding{51}&SwinB&30.0&34.3&9.7&27.1&29.9&11.4&3.9&4.1&2.4&13.6\\
OV2Seg(Ours)&\ding{51}&SwinB&\textbf{37.6}&\textbf{41.1}&\textbf{21.3}&\textbf{33.9}&\textbf{36.7}&\textbf{18.2}&\textbf{4.9}&\textbf{5.3}&\textbf{3.0}&\textbf{17.5}\\
\bottomrule
\end{tabular}}
\caption{Performance comparison on the validation sets of four published datasets: Youtube-VIS19, Youtube-VIS21, BURST, OVIS. The Open in the table indicates whether a method is capable of tackling objects from novel categories following the setting of Open-Vocabulary VIS. The methods above the double horizontal lines are trained on target training videos of each dataset; we only report the overall mean average precision mAP of those methods. The methods below the double horizontal lines are trained on image dataset LVIS and evaluated on the video instance segmentation datasets without fine-tuning; we report both the mAP, mAP${_b}$, and mAP${_n}$ of those methods.\vspace{-4mm}}
\label{tab:vis}
\end{table*}

\noindent \textbf{Class-Dependent vs. Class-Independent Queries.}
OV-DETR~\cite{zang2022opendetr} utilizes $N$ conditional object queries for each category, which we refer to as class-dependent queries $Q^D$.
We compare it with OV2Seg using class-independent queries $Q^I$ and show the results in Fig.~\ref{fig:query_type}.
We see that OV2Seg with class-independent queries $Q^I$ achieves comparable accuracy as with class-dependent queries $Q^D$, while being two orders of magnitude faster.
As shown in Fig.~\ref{fig:query_type} (b), OV2Seg with class-dependent queries becomes significantly slower as the number of categories $|\mathcal{C}|$ increases, and thus is intractable in open-world videos containing objects from large category set.

\begin{table}[t]
\small
\setlength\tabcolsep{4pt}
\centering
\begin{tabular}{c c c c c}
\toprule
\multirow{2}{*}{Methods}&\multicolumn{2}{c}{LV-VIS} &\multicolumn{2}{c}{Youtube-VIS2019}\\
\cmidrule(lr){2-3}
\cmidrule(lr){4-5}
&mAP$_{b}$ &mAP$_{n}$&mAP$_{b}$ &mAP$_{n}$\\
\midrule
Average&16.9&11.3&29.4&10.8\\
Memory Queries&\textbf{17.2}&\textbf{11.9}&\textbf{30.1}&\textbf{11.1}   \\
\bottomrule
\end{tabular}
\vspace{3pt}
\caption{Comparison between two ways to obtain the classification scores. The Memory Queries in the table indicate obtaining classification scores from the memory queries. The Average in the table means directly averaging the per-frame classification scores.\vspace{-5mm}}
\label{tab:class}
\end{table}

\begin{figure}[t]
\vspace{-1mm}
\begin{center}
\includegraphics[width=1\linewidth]{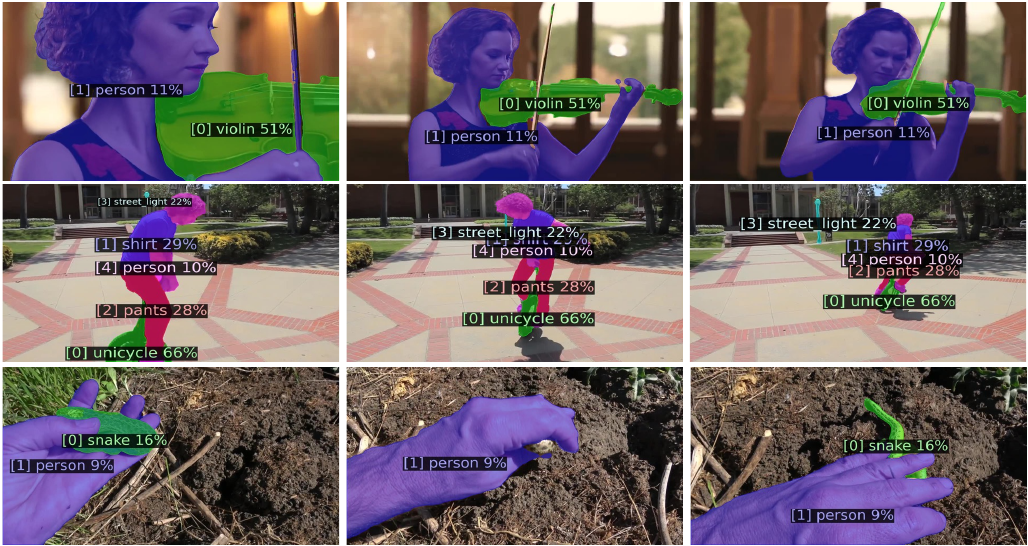}
\end{center}
   \vspace{-5mm}
\caption{Visualizations on LV-VIS dataset.}
\vspace{-5mm}
\label{fig:main_vis}
\end{figure}

\noindent \textbf{Memory Queries.}
We evaluate OV2Seg on LV-VIS validation and Youtube-VIS2019 datasets under different memory update factors $\alpha$ and $s^{\text{obj}}$ in Equation~\ref{eqa:memory}.
As shown in Fig.~\ref{fig:memory_update}, the query association module without memory update, i.e., $\alpha$=0, only achieves 23.7 mAP and 11.9 mAP on Youtube-VIS2019 and LV-VIS validation datasets. 
When $\alpha$ is set to a value between 0.3 and 0.8, OV2Seg achieves stable performance improvements, which shows that the memory update schedule is robust to the factor $\alpha$.
The best performance is obtained by setting $\alpha = 0.7$ and the object score $s^{\text{obj}}$ is adopted, achieving 27.2 mAP and 14.2 mAP on Youtube-VIS2019 and LV-VIS validation dataset.
The memory queries introduce long-term clues to handle occlusion and disappear-reappear cases, which helps to achieve robust object association.
As shown in Fig~\ref{fig:main_vis}, the snack is re-tracked after entire occlusion.

In Tab.~\ref{tab:class}, we compare the performance between the classification by Memory Queries and by Averaging per-frame classification scores.
As shown, the performance of Memory Queries slightly outperforms the naive Average on both LV-VIS and Youtube-VIS2019 datasets.
This is because Memory Queries aggregate video object features based on per-frame object confidence, which constrains the low confidence features and leads to more robust classification with respect to complex cases such as motion blur and occlusion.

\vspace{-3mm}
\subsubsection{Zero-shot Generalization on VIS Datasets.}
We directly evaluate OV2Seg on four video instance segmentation datasets \emph{without finetuning} on the target training videos.
The performance comparison is shown in Tab.~\ref{tab:vis}.
As shown, OV2Seg with ResNet-50 achieves 11.1 mAP$_n$ and 7.3 mAP$_n$ on Youtube-VIS2019 and Youtube-VIS2021 novel categories respectively, surpassing the baseline model Detic-OWTB by 6.6 and 1.5.
Note that without using any training videos of Youtube-VIS, OV2Seg achieves considerable performance compared with close-set fully-supervised competitors such as MaskTrack~\cite{yang2019video} and FEELVOS~\cite{voigtlaender2019feelvos}.
In BURST, OV2Seg achieves 3.7 mAP, outperforming Detic-OWTB by 1.0 mAP in terms of overall mAP.
The BURST dataset is not exhaustively annotated, which makes category-wise evaluation of novel categories statistically unreliable.
On the OVIS dataset with serious object occlusions, OV2Seg achieves 11.2 mAP, comparable with fully-supervised competitors MaskTrack~\cite{yang2019video}, FEELVOS~\cite{yang2019video}, and SipMask~\cite{cao2020sipmask}.
The comparison on OVIS demonstrates that OV2Seg has promising performance in complex occlusion circumstances.
Visualizations are shown in Supplementary Material Sec. \textcolor[rgb]{1,0,0}{C}.\vspace{-2mm}

\section{Conclusion}
We propose a new task of Open-Vocabulary VIS with the goal of segmenting, classifying, and tracking objects of arbitrary categories simultaneously.
To properly evaluate the Open-Vocabulary VIS, we collect a Large-Vocabulary Video Instance Segmentation dataset~(LV-VIS), significantly surpassing the existing video-level segmentation datasets in terms of category sizes and diversity.
Moreover, we propose the first end-to-end Open-Vocabulary VIS model, OV2Seg, that segments, classifies, and tracks objects from arbitrary categories with a Memory-Induced Vision Language Transformer.
Without finetuning on videos, our proposed OV2Seg shows promising zero-shot generalization ability on novel categories unseen during training.


{\small
\bibliographystyle{ieee_fullname}
\bibliography{main}
}

\newpage
\quad
\newpage

\noindent {\fontsize{15}{14}\selectfont \textbf{Appendix}}
\section*{A. Dataset Statistic}

\begin{table*}[!htb]
  \centering
  \scalebox{0.8}{
  \setlength\tabcolsep{5pt}
    \begin{tabular}{llllllllllll}
    \toprule
          &\multicolumn{3}{c}{Basic}  & \multicolumn{4}{c}{Train}& \multicolumn{4}{c}{Val/Test}\\
          \cmidrule(lr){2-4} \cmidrule(lr){5-8} \cmidrule(lr){9-12}  
    Dataset & \multicolumn{1}{l}{Category} & \multicolumn{1}{l}{Length(h)} & \multicolumn{1}{l}{Mask/Frame} & \multicolumn{1}{l}{Video} & \multicolumn{1}{l}{Instance} & \multicolumn{1}{l}{Ann. Frame} & \multicolumn{1}{l}{Mask} & \multicolumn{1}{l}{Video} & \multicolumn{1}{l}{Instance} & \multicolumn{1}{l}{Ann. Frame} & \multicolumn{1}{l}{Mask} \\
    \midrule
    VOT~\cite{kristan2018sixth}       & \multicolumn{1}{l}{-} & 10.7  & 1     & 0     & 0     & 0     & 0     & 62    & 62    & 19,903 & 19,903 \\
    KITTI-MOTS~\cite{voigtlaender2019mots}      & 2     & 39    & 5.4   & 21    & 748   & 8,008 & 38,197 & 28    & 961   & 11,095 & 61,904 \\
    MOTS-Chal.~\cite{voigtlaender2019mots}       & 1     & 34.4  & 10   & 4     & 228   & 2,864 & 26,894 & 4     & 328   & 3,044 & 32,369 \\
    BDD~\cite{yu2020bdd100k}        & 7     & 40    & 11.4 & 154   & 17,838 & 30,745 & 347,442 & 32    & 4,873 & 6,475 & 77,389 \\
    DAVIS17~\cite{pont20172017}   & \multicolumn{1}{l}{-} & 2.9&2.6 & 60    & 144   & 4,219 & 10,238 & 90    & 242   & 6,240 & 16,841 \\
    YT-VOS19~\cite{xu2018youtube}      & \multicolumn{1}{l}{-} & 4.5   & 1.6  & 3741  & 6,459 & 94,400 & 12,918 & 1,048 & 2,115 & 28,825 & 4,310 \\
    UVO~\cite{wang2021unidentified}        & 80$^*$&3&12.3 & 5,641 & 76,627 & 39,174 & 416,001 & 5,587 & 28,271 & 18,966 & 177,153 \\
    \midrule
    YT-VIS19~\cite{yang2019video}      & 40    & 4.5   & 1.7  & 2,238 & 3,774 & 61,845 & 103,424 & 645   & 1,092 & 17,415 & 29,431 \\
    OVIS~\cite{qi2021occluded}      & 25    & 3.2   & 4.7    & 607   & 3,579 & 42,149 & 206,092 & 297   & 1,644 & 20,492 & 89,841 \\
    BURST~\cite{athar2022burst}    & 482   & 28.9  & 3.1      & 500   & 2,645 & 107,144 & 318,200 & 2,414 & 13,444 & 88,569 & 281,957 \\
    \midrule
    LV-VIS~(Ours)    & 1196   & 6.2   & 4.9  & 3,083     & 16,060     & 70,242     & 339,533     & 1,745 & 9,526 & 41,253 & 204,918 \\
    \bottomrule
    \end{tabular}%
}
\caption{Detailed Statistic Comparison between our LV-VIS and other video-level datasets. \textbf{Category}: The number of the overall category set in each dataset. The - in the Category column means the dataset does not provide the category label or take the category into account during evaluation. \textbf{Length}: The total length of videos. \textbf{Mask/Frame}: Average annotated masks per frame.  Statistics for Val/Test on YT-VIS2019, YT-VOS2019, and DAVIS2017 are estimated from the training set, which may not be exact.}
\label{tab:statistic}%
\end{table*}%

\begin{table*}[h]
\setlength\tabcolsep{10pt}
\centering
\begin{tabular}{c c c c c c c}
\toprule
Split&Length(h)&Mask/Frame&Video&Instance&Ann. Frame & Mask\\
\midrule
Train&3.9&4.8&3,083&16,060&70,242&339,533\\
Val&1.1&4.0&838&3,646&19,176&76,916\\
Test&1.3&5.7&908&5749&22,096
&124,834\\
Total&6.2&4.9&4,828&25,588&111,495&544,451\\
\bottomrule
\end{tabular}
\caption{Dataset Split of the LV-VIS.}
\label{tab:split}
\end{table*}

\noindent \textbf{Statistic Comparison.} As shown in Tab.~\ref{tab:statistic}, 
we compare the proposed LV-VIS dataset to existing ones on detailed statistics.

The benchmarks Kitti-MOTS~\cite{voigtlaender2019mots}, BDD~\cite{yu2020bdd100k}, and MOTS-challenge~\cite{voigtlaender2019mots} focus on the automatic driving scenario, which contains long videos of street scenes captured from a driving vehicle or a walking pedestrian.
The benchmarks DAVIS2017~\cite{pont20172017} and Youtube-VOS~\cite{xu2018youtube} mainly focus on semi-supervised video object segmentation~\cite{pont20172017}. The semi-supervised video object segmentation aims to track and segment the objects given the mask of the first frame, which is similar to the single object tracking in VOT~\cite{kristan2018sixth}.
UVO~\cite{wang2021unidentified} is designed for exhaustively segmenting and tracking anything which humans would consider to be "objects" in videos.
Instead of assigning a category label to a specific object, UVO considers all the foreground objects as a single category~(in v1 of UVO, the category labels of objects belonging to MS-COCO are annotated, while the rest objects remain unlabeled, as shown in Fig.~\ref{fig:sm_datasets_composition}).
The category labels are not provided or considered during the evaluation in the above-mentioned datasets.

The Youtube-VIS2019~\cite{yang2019video}, OVIS~\cite{qi2021occluded}, and BURST~\cite{athar2022burst} assign a category label to each annotated object. Therefore, category-wise evaluation is enabled.
However, the Youtube-VIS2019 and OVIS only contain 40 and 25 categories, which is not wide enough for the open-vocabulary evaluation.
The BURST contains a relatively large vocabulary set of 482 categories, but 
81\% of the object instances in BURST are from MS-COCO categories, making it inappropriate for the evaluation of novel categories, as shown in Fig.~\ref{fig:sm_datasets_composition}.
Therefore, BURST only considers the 80 common categories in MS-COCO in the category-wise evaluation and mainly follows the evaluation protocols of class-agnostic multiple object tracking.

By contrast, our LV-VIS dataset not only contains a large vocabulary set of 1,196 categories but has a diverse category and object instances distribution, as shown in Fig.~\ref{fig:sm_datasets_composition}.
Specifically, 94\% categories in LV-VIS are disjointed with categories in MS-COCO, while 46\% are disjointed with frequent/common categories in LVIS.
Moreover, OV-VIS contains 4,828 videos and 544,451 annotated masks for evaluation, which is much larger than most of the validation/test sets in existing datasets.
The dataset split detail of LV-VIS is shown in Tab.~\ref{tab:split}. As shown, we divide LV-VIS into a training set, a validation set, and a test set, where the test set is relatively more complex than the validation set.

\begin{figure*}[t!]
\begin{center}
\includegraphics[width=1\linewidth]{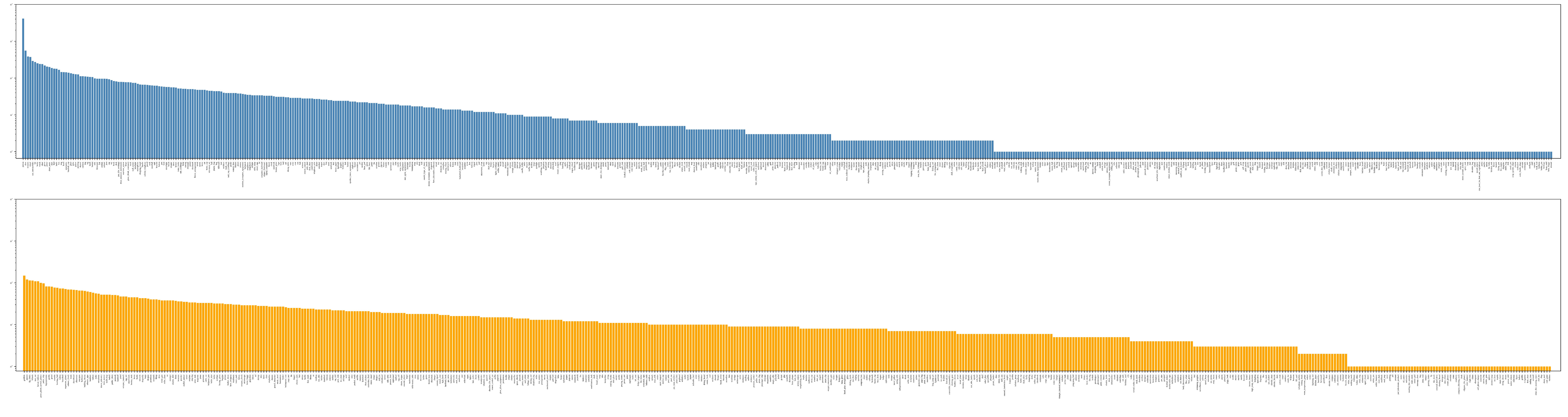}\vspace{-3mm}
\end{center}
\caption{The instance per category on LV-VIS dataset. The blue bars indicate the base categories~(frequent/common categories in LVIS) and the orange bars indicate the novel categories~(disjointed with base categories).
}
\label{fig1:instance}
\end{figure*}

\begin{figure*}[ht]
\vspace{-1mm}
\begin{center}
\includegraphics[width=0.7\linewidth]{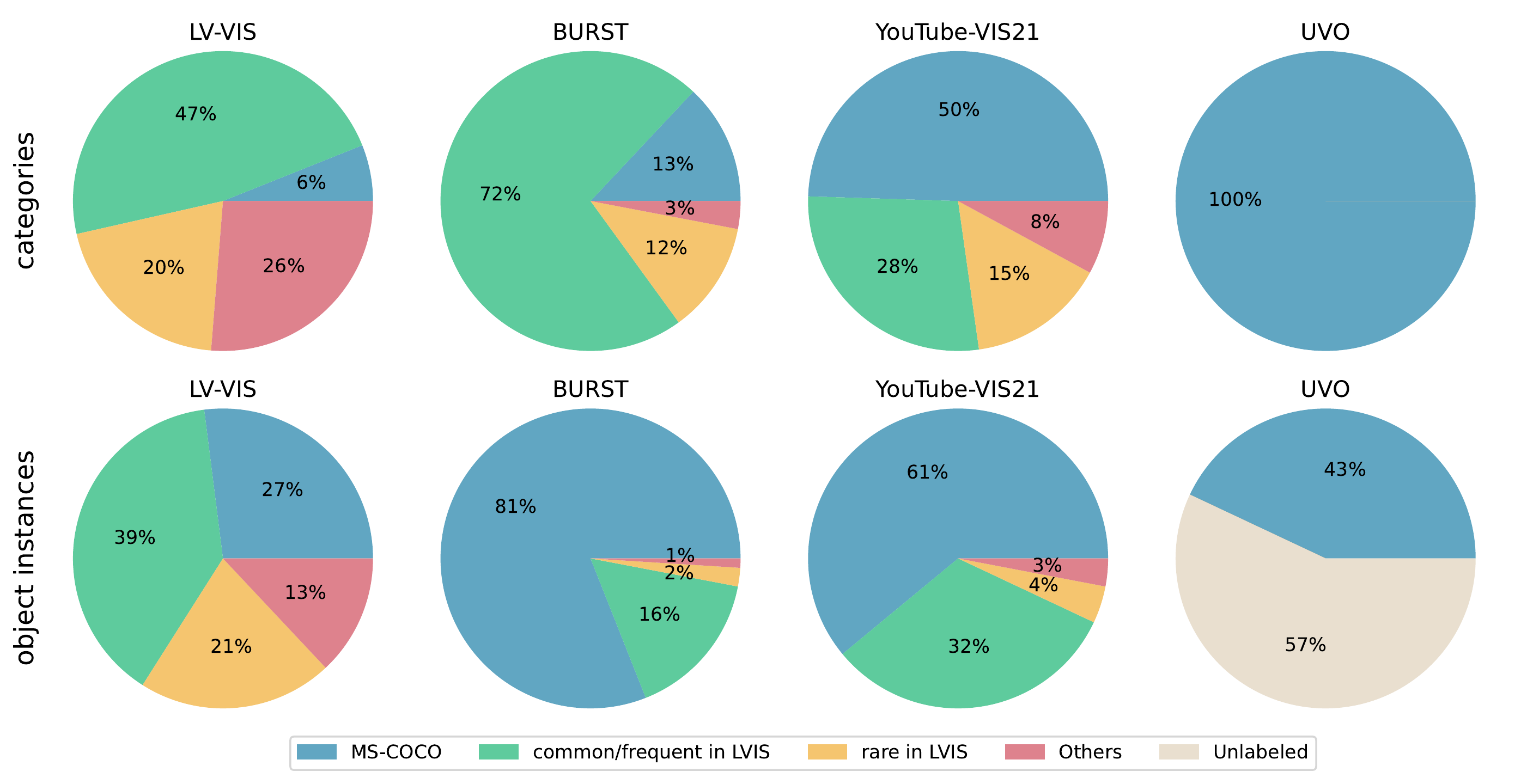}
\end{center}
   \vspace{-5mm}
\caption{Category and object instance distributions.  To avoid overlap, categories from MS-COCO are excluded in common/frequent LVIS to draw the figures. A significant proportion of categories and objects in the LV-VIS are distinct from the commonly used datasets.}
\vspace{1mm}
\label{fig:sm_datasets_composition}
\end{figure*}

\begin{figure*}[ht]
\vspace{-1mm}
\begin{center}
\includegraphics[width=1.0\linewidth]{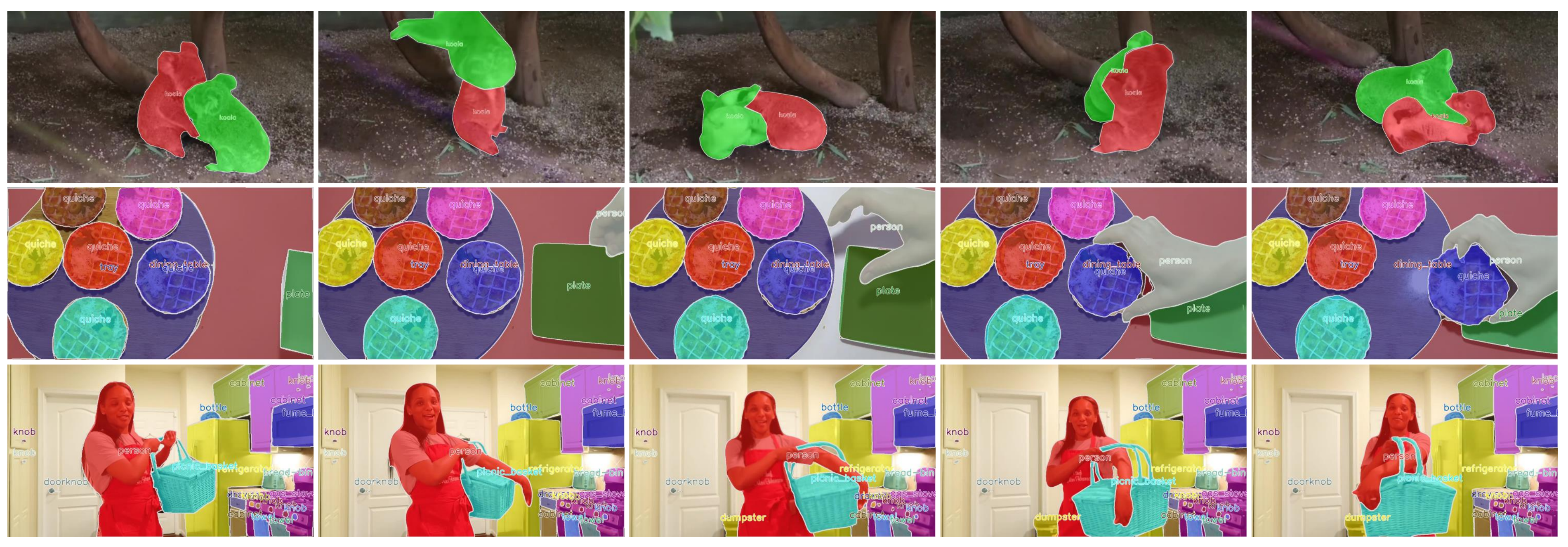}
\end{center}
   \vspace{-5mm}
\caption{Sample videos in LV-VIS.}
\vspace{-1mm}
\label{fig:anno_vis}
\end{figure*}

\vspace{5pt}
\noindent \textbf{Category Partition.}
In this section, we first illustrate how we select the 1,196 object categories.
We first include all 1,203 categories in LVIS. Then we manually select novel categories in ImageNet21K to cover diverse types of categories, such as animals, plants, vehicles, tools, clothing, food, {\em etc.} In this way, we get 1,612 category candidates. 
As for categories with multiple meanings, we manually add additional descriptions like LVIS dataset, e.g.,  date\_(fruit), triangle\_(musical\_instrument). 
There exist two types of relationships between each of the category pairs: disjoint and non-disjoint. 
Non-disjoint category pairs could be in partially overlapping, parent-child, or equivalent relationships, which means a single object could have multiple valid category labels. 
Therefore, we first manually merge object categories with the same semantics, ensuring that there are no mutually equivalent category pairs.
In this way, we collect 1,196 well-defined mutually different object categories.
Then we construct a parent-child relationship tree among the collected categories. 
For instance, "race car" is a child of "car".
Finally, we revise the annotation based on the defined parent-child relationship tree, ensuring that each object in videos is exhaustively annotated as all corresponding object categories.
For instance, a "race car" is also assigned to the corresponding parent label "car".
With the aforementioned pipeline, we address the annotation issues of equivalent categories, parent-child categories, and partially overlapping categories.
The category partition and the number of instances per category of the LV-VIS dataset are shown in Tab.~\ref{fig1:instance}, where the 1,196 categories are divided into 641 base categories and 555 novel categories. All the base categories in LV-VIS are inherited from frequent/common categories in LVIS~\cite{gupta2019lvis}. 
While conducting the evaluation, the categories in Youtube-VIS2019 are divided into 33 base categories and 7 novel categories.
The categories in Youtube-VIS2021 are divided into 34 base categories and 6 novel categories.
We show the category partitions of Youtube-VIS2019 and Youtube-VIS2021~\cite{yang2019video} in Tab.~\ref{youtube}.

\textbf{Annotation Details.} We develop a video segmentation annotation platform based on Labelme, which is released here \textcolor[rgb]{0,0,1}{https://github.com/haochenheheda/segment-anything-annotator}. 
We first manually annotate all object masks in the first frame by polygons and propagate the object masks to the second frame with STCN~\cite{cheng2021rethinking}. 
After that, we correct the propagated masks, add masks for newly appeared objects, and then repeat the propagation to the next frame. 
We manually recognize and assign category names to each annotated mask sequence. 
Finally, we include cross-revision to ensure the annotation quality.



\section*{B. Visualizations of Annotated Frames}
Examples of annotated videos in LV-VIS are shown in Fig.~\ref{fig:anno_vis}.

\section*{C. Visualization and Failure Case Analysis}
We demonstrate the result of OV2Seg on our proposed LV-VIS, Youtube-VIS~\cite{yang2019video} and OVIS~\cite{qi2021occluded} in Fig.~\ref{fig1:result}.
OV2Seg shows a strong generalization ability on those video instance segmentation datasets, even in some hard cases, to be specific, (b) large perspective change, (c) blurry video, (f) long video, (a, g) occlusion, and (g) a large number of objects. 

The failure cases are demonstrated in Fig~\ref{fig1:result} (h)-(i).
The major failure case is category confliction,
which means the classification of objects from novel categories is usually dominated by their visually similar base categories.
To be specific, as demonstrated in Fig~\ref{fig1:result} (i), The "wolf" in the figure is recognized as a "dog" because of the apparent similarity.
As the objects of "dog" are shown in the training set while the objects from "wolf" are not, the model learns better alignments between the object embedding and the word embedding of "dog," which makes the model tend to recognize an object as "dog" instead of a "wolf."
We consider the category confliction as a fundamental challenge for all the open-vocabulary tasks, which could be improved by including a large vocabulary set during training or some training protocol to transfer the information from the image domain, such as knowledge distillation or self-training.
Another failure case is the miss segmentation of some common categories, such as the "person" in Fig.~\ref{fig1:result} (b), (h), (i).
This is because the LVIS is not a densely annotated dataset. Only a part of the objects are annotated, especially for the most common objects such as "person."
Specifically, LVIS only annotated 13,439 "persons" out of 262,465 (annotated in MS-COCO~\cite{lin2014microsoft}) in total.
Therefore most of the "person" objects in the training set are regarded as background, which leads to a low recall of the person category.
This could be relieved by involving some semi-supervised training methods or combining the LVIS with the MS-COCO~\cite{lin2014microsoft} dataset to fill the miss annotations of the common categories in LVIS.
We hope to inspire future works by giving analyses of the failure cases.

\newpage

\begin{table*}[h]
\begin{tabular}{cclllllll}
\toprule[1pt]
Datasets                         & Type                      & \multicolumn{7}{c}{Categories}                                                            \\ 
\midrule[0.8pt]
\multirow{6}{*}{Youtube-VIS2019} & \multirow{5}{*}{Base}     & airplane       & bear   & boat     & cat        & cow         & deer        & dog           \\
                                 &                           & duck         & eagle  & elephant & fish       & frog        & giant panda & giraffe       \\
                                 &                           & horse        & lizard & monkey   & motorbike  & mouse       & owl         & parrot        \\
                                 &                           & person       & rabbit & shark    & skateboard & snowboard   & surfboard   & tennis racket \\
                                 &                           & tiger        & train  & truck    & turtle     & zebra       &             &               \\ \cline{2-9} 
                                 & Novel                     & earless seal & fox    & leopard  & snake      & ape         & hand        & sedan         \\ \midrule[0.7pt]
\multirow{6}{*}{Youtube-VIS2021} & \multirow{5}{*}{Base}     & airplane     & bear   & boat     & car        & cat         & cow         & deer          \\
                                 &                           & dog          & duck   & eagle    & elephant   & fish        & frog        & giant panda   \\
                                 &                           & giraffe      & house  & lizard   & monkey     & motorbike   & mouse       & owl           \\
                                 &                           & parrot       & person & rabbit   & shark      & skateboard  & snowboard   & surfboard     \\
                                 &                           & tennis       & tiger  & train    & truck      & turtle      & zebra       &               \\ \cline{2-9} 
                                 & \multicolumn{1}{l}{Novel} & earless seal & fox    & leopard  & snake      & flying dsic & whale       &               \\ 
                                 \bottomrule[1pt]
\end{tabular}
\caption{Base and novel categories in Youtube-VIS2019 and Youtube-VIS2021 datasets.}
\label{youtube}
\end{table*}

\begin{figure*}[!htb]
\begin{center}
\includegraphics[width=0.92\linewidth]{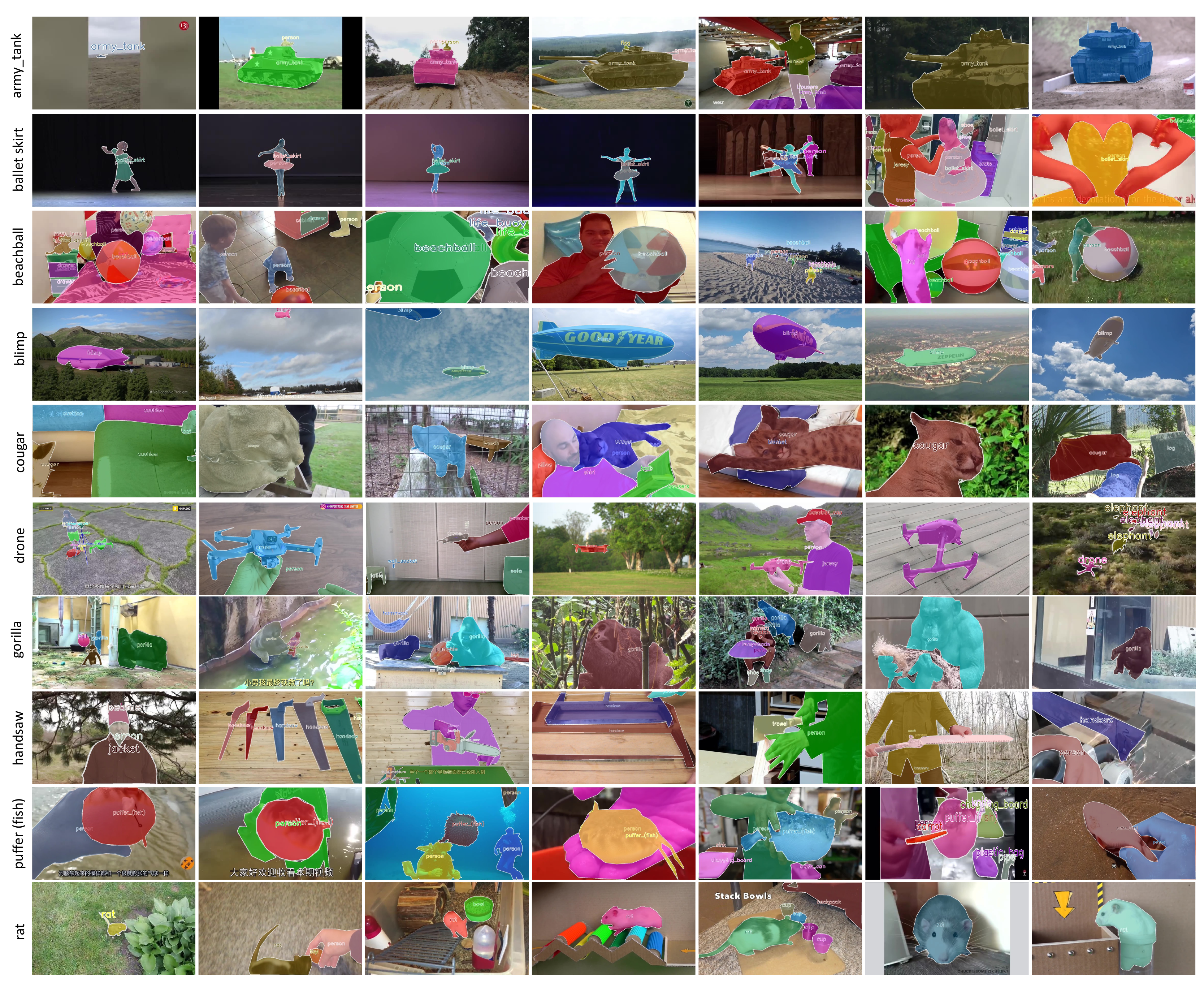}
\end{center}
   \vspace{-5mm}
\caption{A screenshot of annotated frames in LV-VIS. The full videos and annotations will be released upon publication.
\vspace{-2mm}}
\label{fig1:anno}
\end{figure*}

\newpage

\vspace*{\fill}
\newpage

\begin{figure*}[b]
\begin{center}
\includegraphics[width=1\linewidth]{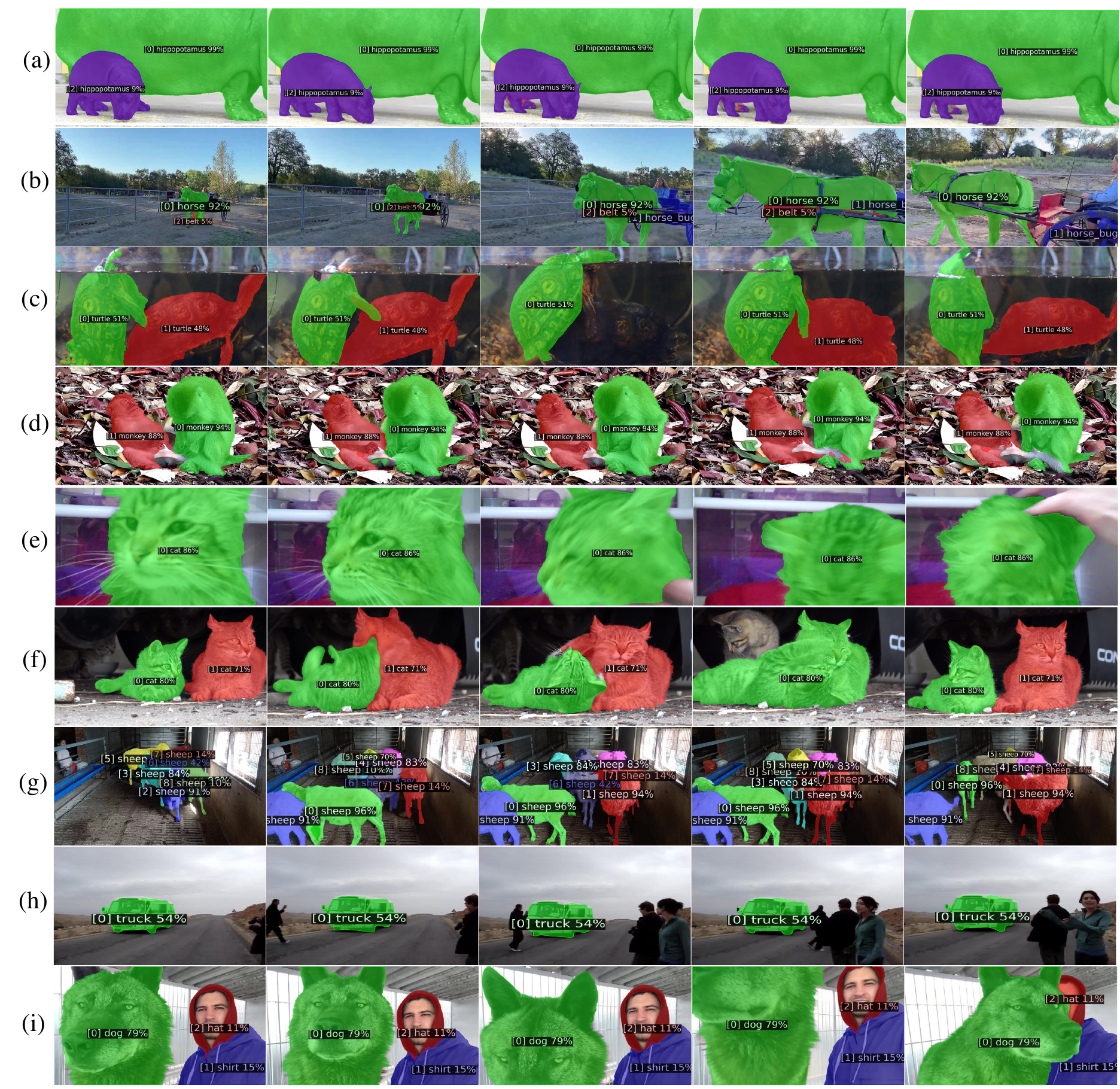}
\end{center}
   \vspace{-5mm}
\caption{Predictions of OV2Seg on Video Instance Segmentation datasets. Figures (a), (b), and (i) are from the LV-VIS dataset. Figures (c), (d), (e), and (h) are from Youtube-VIS2019/2021 datasets. Figures (f) and (g) are from the OVIS dataset. Figures (h) and (i) are failure cases.
\vspace{-16mm}}
\label{fig1:result}
\end{figure*}
\end{document}